\PassOptionsToPackage{dvipsnames,table}{xcolor}

\documentclass[10pt,twocolumn,letterpaper]{article}
\usepackage[accsupp]{axessibility}
\usepackage{iccv}
\definecolor{iccvblue}{rgb}{0.21,0.49,0.74}
\usepackage[pagebackref,breaklinks,colorlinks,allcolors=iccvblue]{hyperref}
\usepackage{algorithmicx}  
\usepackage{algpseudocode}
\usepackage{algorithm}   
\usepackage{tabularx}
\usepackage{makecell} 
\usepackage{booktabs} 
\usepackage{colortbl}
\usepackage{tcolorbox}

\let\oldtwocolumn\twocolumn
\renewcommand\twocolumn[1][]{
    \oldtwocolumn[{#1}{
        \centering
        \vspace{-20pt}
        \includegraphics[width=\textwidth]{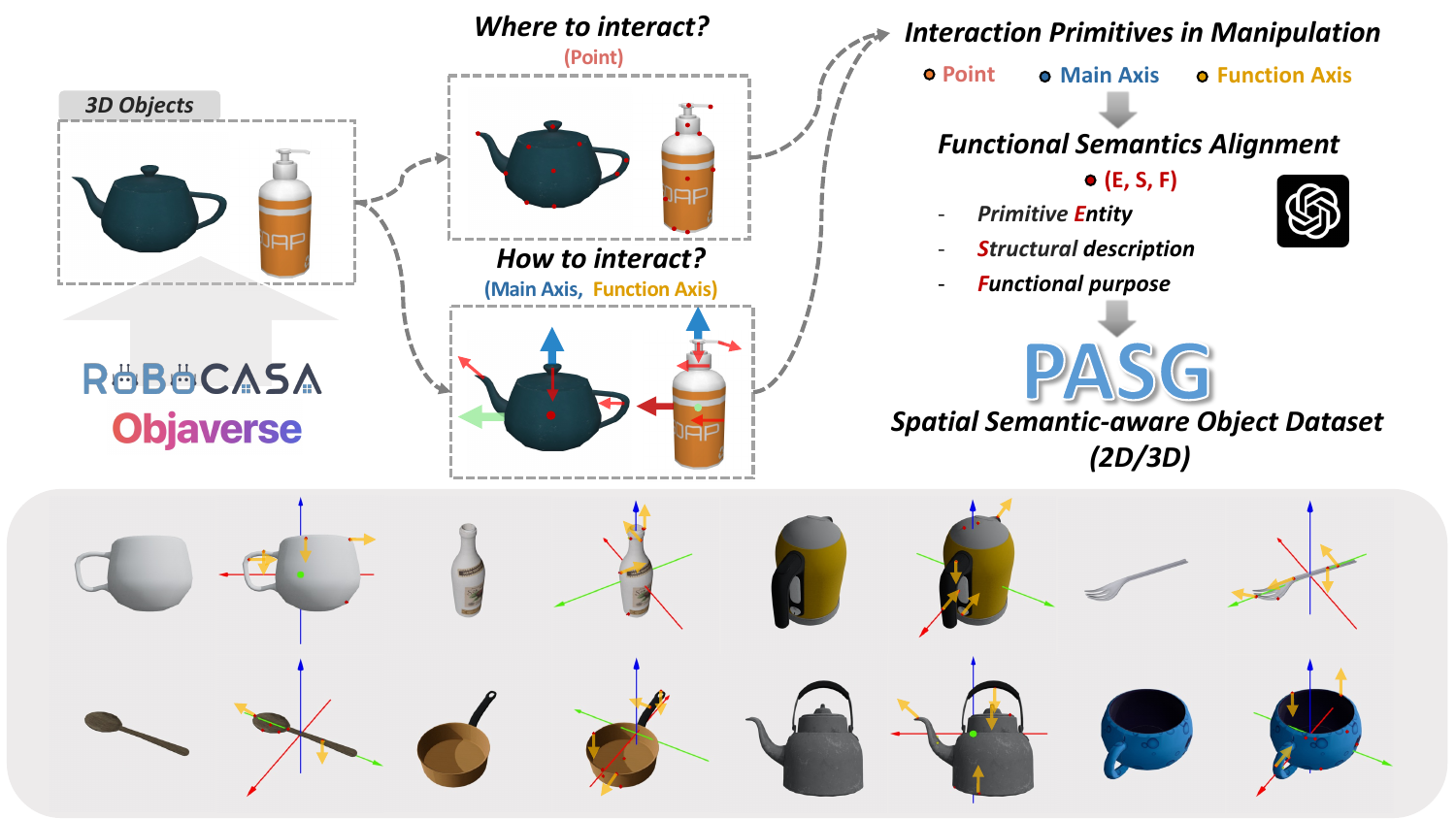}  
        \vspace{-15pt}
        \captionof{figure}{We propose PASG, an automated object-centric spatial-semantic enhancement framework for robotic manipulation. By formalizing interaction primitives and establishing semantic-geometric correspondences, our approach achieves structural coupling between low-level spatial primitives and high-level functional semantics, enabling joint enhancement of manipulation reasoning and semantic-aware 2D/3D object dataset generation.}      
        \vspace{5pt}
        \label{fig:teaser}
    }]
}

\title{PASG: A Closed-Loop Framework for Automated Geometric Primitive Extraction and Semantic Anchoring in Robotic Manipulation}

\author{\textbf{Zhihao Zhu}\thanks{Equal Contribution. 
}\hspace{4pt} \quad \textbf{Yifan Zheng}$^{*}$ 
\quad \textbf{Siyu Pan}$^{*}$  \quad \textbf{Yaohui Jin}\thanks{Yao Mu and Yaohui Jin are the corresponding authors 
}\hspace{4pt} \quad 
\bf{Yao Mu}$^{\dag}$
 \\
MoE key Lab of Artificial Intelligence, AI Institute, Shanghai Jiao Tong University \\
{\small \texttt{\{zzh2021, yifanzheng, pansiyu0327, jinyh, muyao\}@sjtu.edu.cn}}
}

\begin{document}
\maketitle
\begin{abstract}
The fragmentation between high-level task semantics and low-level geometric features remains a persistent challenge in robotic manipulation. While vision-language models (VLMs) have shown promise in generating affordance-aware visual representations, the lack of semantic grounding in canonical spaces and reliance on manual annotations severely limit their ability to capture dynamic semantic-affordance relationships. To address these, we propose \textbf{P}rimitive-\textbf{A}ware \textbf{S}emantic \textbf{G}rounding (\textbf{PASG}), a closed-loop framework that introduces: (1) Automatic primitive extraction through geometric feature aggregation, enabling cross-category detection of keypoints and axes; (2) VLM-driven semantic anchoring that dynamically couples geometric primitives with functional affordances and task-relevant description; (3) A spatial-semantic reasoning benchmark and a fine-tuned VLM (Qwen2.5VL-PA). We demonstrate PASG’s effectiveness in practical robotic manipulation tasks across diverse scenarios, achieving performance comparable to manual annotations. PASG achieves a finer-grained semantic-affordance understanding of objects, establishing a unified paradigm for bridging geometric primitives with task semantics in robotic manipulation. 
\end{abstract}    
\section{Introduction}
\label{sec:introduction}

Developing generalizable robotic manipulation in unstructured environments remains challenging due to the semantic asymmetry between low-level interaction primitives (points/axes) and high-level task planning. As large language models (LLMs)\citep{achiam2023gpt, touvron2023llama} and vision-language models (VLMs)\citep{radford2021learningtransferablevisualmodels, alayrac2022flamingo, li2023blip, zhang2024vision} demonstrate promise in semantic reasoning and commonsense knowledge, researchers have attempted to integrate these models into robotic manipulation\citep{gao2024physically, Chen_2024_CVPR, duan2024manipulateanythingautomatingrealworldrobots, liu2024moka}. However, these approaches primarily focus on high-level task decomposition and planning\citep{singh2023progprompt, kannan2024smart, wang2024llm, obata2024lip, zhou2024isr}. In contrast, the capacity of semantic reasoning within 3D spatial primitives remains underdeveloped. This limitation stems from insufficient semantic understanding of object canonical spaces—for instance, manually annotated "handle centers" for teapots lack contextual semantics (such as functional descriptions and usage scenarios), leading to inaccurate spatial constraint reasoning—revealing the inherent fragility of strategies that directly map task semantics to canonical spaces devoid of semantic context.

To endow robots with spatial primitive understanding, current approaches typically fine-tune VLMs on large-scale manipulation demonstrations to enhance spatial semantic reasoning. However, such methods depend on manually annotated geometric primitives (e.g., keypoints, axes), which leads to high annotation costs and inherently limits generalizability. Recent pioneering work leverages pre-trained large vision models (LVM) \citep{alayrac2022flamingo, radford2021clip} to detect interaction features, followed by VLM-based semantic filtering to identify task-relevant primitives \citep{, huang2023voxposercomposable3dvalue}. Nevertheless, such frameworks exhibit two systemic weaknesses: (1) Automated detection methods (e.g., SAM \citep{Kirillov_2023_ICCV}, DINOV2 \citep{oquab2024dinov2learningrobustvisual}) lack verification mechanisms, propagating errors from undetected or misaligned primitives and drastically degrade success rates; (2) Incomplete canonical space definitions - overlooking essential orientation features of object like main axes while only include keypoints and directions - result in manipulation failures during grasping or transportation tasks~\cite{yu2024maniposecomprehensivebenchmarkposeaware}. These limitations underscore the need for a unified framework that integrates automated primitive extraction with semantic-task contextualization.

To address these challenges, as shown in Fig~\ref{fig:teaser}, we propose PASG, a closed-loop framework establishing the mapping between spatial primitives and functional semantics. It offers several key innovations: First, our geometry-aware feature extraction module automatically detects interaction primitives (keypoints, directions, and principal axes) through visual foundation models (VFMs) integration with geometric topology analysis, without any manual annotation. Second, our dynamic semantic anchoring mechanism employs VLMs to contextualize primitives with multi-granularity semantics - from low-level descriptions ("edge of the neck") to high-level intents ("Crucial for aligning the bottle during pouring or filling.") - while implementing self-corrective feature extraction loops. Third, we validate the practicality of PASG through extensive manipulation experiments conducted on diverse tasks within a simulated environment, demonstrating competitive or superior performance compared to human annotations. Lastly, we provide Robocasa-PA, an extensive benchmark that supports scalable, object-based evaluations of functional primitive understanding in manipulation scenarios, and develop Qwen2.5VL-PA through parameter-efficient LoRA fine-tuning, achieving 77.8\% overall accuracy (+33.9\% absolute improvement) with minimal cross-domain variance.

Our contributions are as follows:
\begin{itemize}
\item We propose a novel framework that automatically annotates hierarchical semantics for object interaction primitives, bridging the gap between low-level geometric features and high-level task semantics.
\item We introduce Robocasa-PA, featuring 8,343 validated visual questions across three task evaluations, providing the first benchmark for assessing functional primitive understanding in manipulation.
\item We demonstrate PASG's effectiveness in real-world manipulation scenarios, achieving competitive performance relative to human annotations and enhancing diversity and flexibility in grasp and interaction primitives.
\end{itemize}

\section{Related Work}
\label{sec:relatedwork}
\begin{table*}[ht!]
\centering
\label{tab:framework_comparison}
\renewcommand{\arraystretch}{1}
\setlength{\tabcolsep}{4pt}
\begin{tabularx}{\textwidth}{l>{\centering\arraybackslash}X>{\centering\arraybackslash}X>{\centering\arraybackslash}X>{\centering\arraybackslash}X} 
\toprule
\textbf{Method} & \textbf{Geometric Primitives} & \textbf{Primitive Extraction} & \textbf{Semantic Coupling} & \textbf{Adaptive Refinement\textsuperscript{†}} \\
\midrule

\textbf{ReKep \citep{huang2024rekep}} 
& Keypoints 
& \makecell{VFM + VLM Detection \\ \textit{(Task-Level)}} 
& No
& No
\\

\textbf{CoPa \citep{huang2024copa}} 
& \makecell{Keypoints, Function \\ Axes} 
& \makecell{VFM + VLM Detection \\ \textit{(Task-Level)}} 
& No
& No
\\

\textbf{OmniManip \citep{pan2025omnimanipgeneralroboticmanipulation}} 
& \makecell{Keypoints, Main \\Axes} 
& \makecell{VFM + VLM Detection \\ \textit{(Task-Level)}} 
& No
& Yes
\\

\textbf{FUNCTO \citep{tang2025functofunctioncentriconeshotimitation}} 
& \makecell{Keypoints, Function \\ Axes} 
& \makecell{VFM + VLM Detection \\ \textit{(Task-Level)}} 
& Predefined Semantics
& No
\\

\textbf{Robotwin \citep{mu2024robotwindualarmrobotbenchmark}} 
& \makecell{Keypoints, Function \\Axes, Main Axes} 
& \makecell{Human-Annotated \\ \textit{(Object-Level)}} 
& Predefined Semantics 
& No
\\

\textbf{SoFar \citep{qi2025sofarlanguagegroundedorientationbridges}} 
& Main Axes 
& \makecell{Predefined Orientation \\ \textit{(Object-Level)}} 
& \makecell{Automatic Spatial \\ Semantic Anchoring}
& No
\\

\rowcolor{gray!20} 
\textbf{PASG (Ours)} 
& \makecell{Keypoints, Function \\Axes, Main Axes} 
& \makecell{VFM + VLM Detection \\ \textit{(Object-Level)}} 
& \makecell{Automatic Spatial \\ Semantic Anchoring}
& Yes
\\

\bottomrule
\end{tabularx}
\caption{Normative interaction primitive and semantic coupling across different frameworks in robotic manipulation tasks: PASG as the first automated closed-loop framework with primitive extraction, semantic anchoring, and self-refinement. \textsuperscript{†} Adaptive refinement refers to the mechanism that self-corrects erroneous or omitted geometric primitives. OmniManip employs computational constraint optimization and scene rendering for VLM validation, while our method directly detects annotation-primitive misalignment for efficient self-correction.}
\vspace{-8pt}
\end{table*}

\subsection{Language-Grounded Manipulation}
Natural language has emerged as a critical interface for robotic manipulation. Existing approaches fall into two categories: (1) End-to-end models: Cross-modal Transformers \citep{frank2021visionandlanguagevisionforlanguagecrossmodalinfluence, 10163247, zhang2023cmx, messina2021fine}, and Vision-Language-Action (VLA) models trained on large-scale robotic datasets (e.g., RT-X, Open X-Embodiment) \citep{lu2019vilbert, brohan2022rt, brohan2023rt, o2024open}, unify perception \citep{huang2023language, liu2024uni3dllmunifyingpointcloud}, planning \citep{fan2024vision, zhen20243dvla3dvisionlanguageactiongenerative}, and action cross-modally through latent space alignment \citep{maiorca2023latent, he2024multi, wang2022optimizing}, demonstrating strong generalization. However, their dependence on domain-specific robotic data introduces scalability bottlenecks. (2) Decoupled Language-Planning Architectures: This approache separates low-level motion control from high-level task reasoning, leveraging VLMs for instruction-based subgoal decomposition \citep{garrett2021integrated, guo2023recent, ding2023task, wang2024llmˆ, chen2024autotamp} and formulate constraint optimization problems based on geometric primitives \citep{huang2024rekep, huang2024copa, mu2024robotwindualarmrobotbenchmark}. This method effectively harnesses VLMs' multimodal semantic reasoning capabilities to enhance interpretability. However, these methods suffer from coarse coupling mechanisms that lead to primitive-semantic misalignment. Contemporaneous work SOFAR addresses this limitation by proposing direction-aware spatial understanding module PointSO, which connects geometric reasoning with functional semantics \citep{qi2025sofarlanguagegroundedorientationbridges}. Unlike SOFAR's predefined directional priors, PASG framework focuses on fine-grained keypoints and functional vectors to construct hierarchical semantic anchoring, achieving deeper integration between task semantics and spatial primitives.

\subsection{Spatial Reasoning for Manipulation} 
Spatial reasoning in manipulation involves inferring interaction constraints from object's spatial primitives to guide robot actions. Keypoints are widely adopted for constructing spatial constraints due to their semantic interpretability and ease of representation \citep{9617128, 10.1007/978-3-031-20086-1_31, 7989233}. However, manual annotation of keypoints limits scalability. Recent advances address this by leveraging visual prompts (scene images) fed directly into vision-language models (VLMs) to autonomously identify task-relevant points, achieving zero-shot consistency \citep{tang2025functofunctioncentriconeshotimitation, abdelreheem2023satrzeroshotsemanticsegmentation, kim2024partstad2dto3dsegmentationtask}. Nevertheless, the discrete nature of keypoints and detection instability degrade reasoning performance while failing to provide effective guidance for end-effector pose optimization. To address this, researchers integrate 6D pose estimation with positional awareness, significantly enhancing end-effector pose stability \citep{pan2025omnimanipgeneralroboticmanipulation, Wen_2024_CVPR, 10611464}. However, traditional orientation representations suffer from sparsity and static template-based definitions that lack semantic foundations, limiting adaptability in open-world manipulation. PASG resolves these by augmenting geometric primitives with function-aware directional vectors (e.g., "pressing direction" for buttons, "grasping direction" for handles), enabling flexible yet stable directional constraints. Through automated extraction of primitives using VFMs and hierarchical semantic alignment with VLMs, PASG achieves closed-loop optimization of spatial reasoning paradigms in open-world scenarios.

\section{Method}
\label{sec:method}

In this section, we explore the following research questions: (1) How to define spatial geometric primitives for objects in manipulation tasks? (2) How to automatically extract geometric primitives from objects? (3) How does PASG achieve dynamic alignment between geometric primitives and task semantics? (4) How to leverage PASG to enhance reasoning in manipulation tasks?

\begin{figure*}[t]
    \centering
    \includegraphics[width=\linewidth]{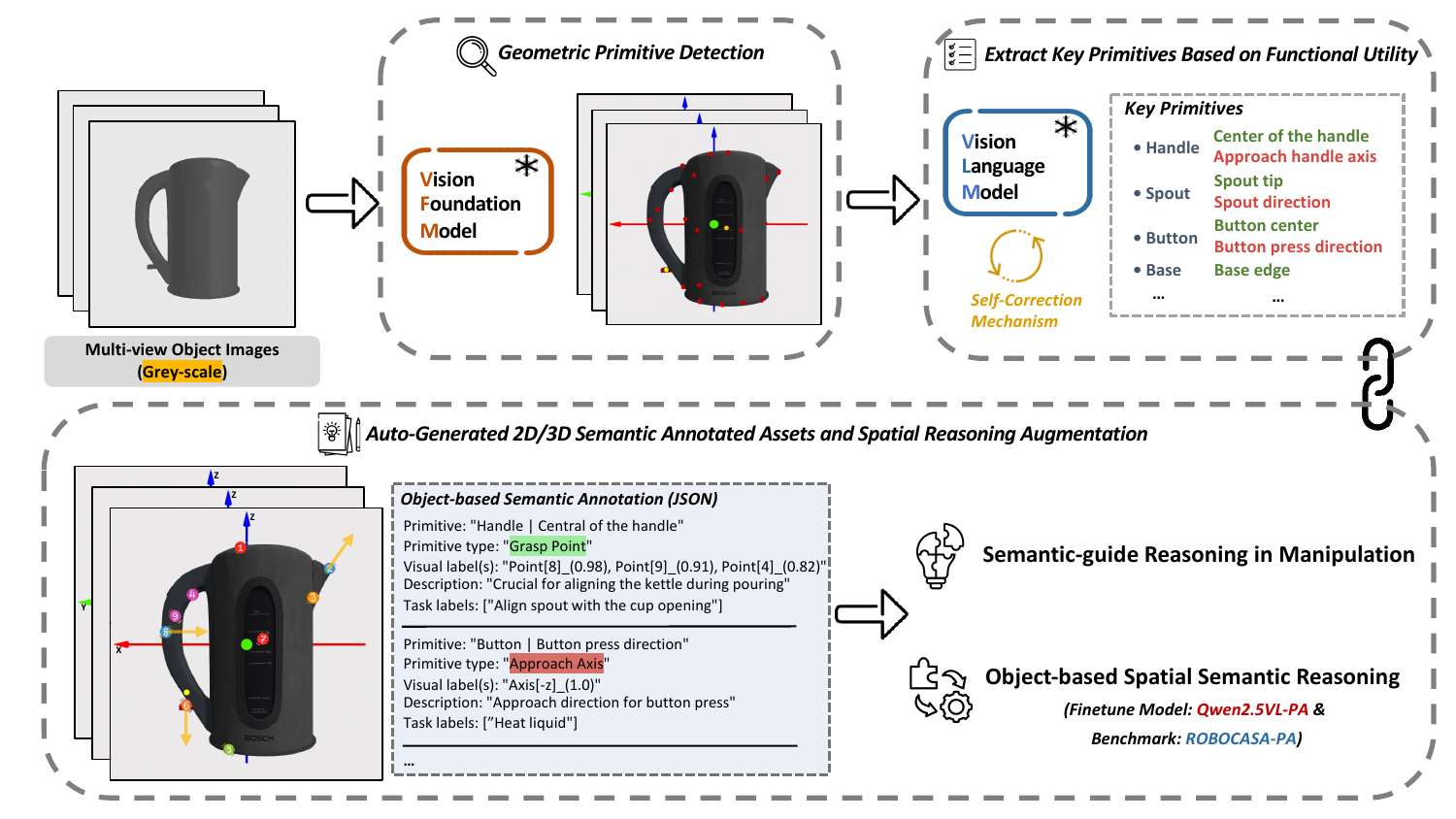} 
    \vspace{-20pt}
    \caption{Overview of PASG} 
    \label{fig:overview}
    \vspace{-15pt}
\end{figure*}

\subsection{Semantic Primitives in Robotic Manipulation}

In robotic manipulation tasks, spatial primitives of objects serve as fundamental building blocks for planning and executing actions. Traditionally, these primitives are defined by geometric entities \( E \) (e.g., points, axes, orientations). However, beyond pure geometry, each primitive often carries both semantic \( S \) and functional information \( F \). For instance, a cup's handle represents not merely a spatial protrusion but provides an affordance for grasping while implicitly suggesting proper manipulation strategies. By augmenting geometric primitives with such attributes, we can better capture an object's intended usage and operational constraints. To formally integrate these aspects, we define an \textit{interaction primitive} as a triplet \( (E, S, F) \) that combines geometric, structural, and functional properties:

\begin{itemize}
    \item \( E \) -- Geometric primitive entity
    \item \( S \) -- Structural description of the object component
    \item \( F \) -- Functional role in manipulation tasks
\end{itemize}

To further integrate operational task semantics, we categorize interaction primitives into two functionally distinct classes based on manipulation requirements: \textit{point-based} (\( \mathcal{P} \)) and \textit{axis-based} (\( \mathcal{A} \)) primitives.

\noindent\textbf{Point Interaction Primitives (\( \mathcal{P} \))} denote specific object locations critical for manipulation.
\begin{itemize}
    \item \textbf{Anchor Point} (\( p_a \)): A reference position that determines how an object should be placed or aligned in the workspace. (e.g. the spout tip over a cup for pouring)
    \item \textbf{Grasp Point} (\( p_g \)): A location on the object optimized for a secure hold by a robot’s end-effector. (e.g. the center of a mug’s handle)
    \item \textbf{Actuation Point} (\( p_{act} \)): The specific spot that triggers a mechanism or function when pressed or manipulated.(e.g. the power button on a microwave)
\end{itemize}

\noindent\textbf{Axis Interaction Primitives (\( \mathcal{A} \))} encode directional information derived from geometric properties and functional requirements. These specify object orientation and motion constraints:
\begin{itemize}
    \item \textbf{Primary Axis} (\( a_p \)) The principal orientation axis of the object, usually dictated by its geometry or symmetry.(e.g. vertical primary axis \( z \) in a teapot)
    \item \textbf{Functional Axis} (\( a_f \)): The axis aligned with the object’s intended action or function – essentially the direction along which the object exerts its effect. (e.g. the hammerhead direction in a hammer)
    \item \textbf{Approach Axis} (\( a_{app} \)): The direction from which a robot’s end-effector should approach the object to interact with a specific point. (e.g. approaching a handle from the side along the handle’s orientation)
\end{itemize}

\subsection{Geometry Primitive Extraction}
\label{sec:geom_primitive_extraction}
To obtain accurate spatial primitive detection, we employ a VFM for fine-grained region segmentation, followed by geometric-topological processing for hierarchical keypoint detection and filtering.

\noindent\textbf{VFM-Based Region Segmentation} We utilize a pre-trained VFM (Semantic SAM \citep{li2023semantic}) for fine-grained semantic segmentation. To enable this, we first acquire multi-view RGB images (\( \mathcal{I} = \{I_1,...,I_n\} \)) from the object's 3D mesh data, which are then resized and fed into the segmentation model. Inspired by SoM \citep{yang2023setofmarkpromptingunleashesextraordinary}, we preprocess the obtained segmentation masks (\( \mathcal{M} = \{M_1,...,M_n\} \))  through connected component analysis to extract the largest foreground region while eliminating small areas and background noise. 

\noindent\textbf{Keypoint Extraction} For geometric keypoint (\( \mathcal{K}_{\text{raw}} \)) detection, we extract representative geometric positions including centers  (\( \mathcal{C} = \{c_1,...,c_n\} \)) and corner feature points (\( \mathcal{F} = \{f_1,...,f_n\} \)) in segmentation masks \( M \). The centroid (\( c \)) is calculated through mask moments. Corner detection employs polygon approximation (cv2.approxPolyDP()) and curvature analysis. Additionally, Principal Component Analysis (PCA) calculates two orthogonal axes from the mask, where their intersections with the boundary are extracted as supplementary feature points.

\noindent\textbf{Keypoint Filtering} To reduce redundancy in detected keypoints (\( \mathcal{K}_{\text{raw}} \)), we implement a two-stage filtering mechanism: DBSCAN clustering removes locally dense points, while optimized farthest point sampling globally selects distinctive features, yielding final keypoints (\( \mathcal{K}_{\text{filter}}  \)) including center point set (\( \mathcal{C} \)) and feature points set (\( \mathcal{F} \) ).

\noindent\textbf{Principal Axis Calibration} For standardized axis representation, most 3D objects in datasets provide pre-aligned main axes. We rectify significant deviations by defining the Z-axis as the line connecting geometric centroids of top/bottom view masks. Orthogonal X/Y axes are subsequently generated. To ensure cross-view consistency, we enforce color standardization for axis visualization across different objects and viewpoints.

\subsection{Task-Oriented Semantic Annotation}
\label{sec:task_semantic_annotation}
To generate task-oriented primitive annotations, we establish a mapping from task intentions to spatial primitives through three core stages: task-driven semantic primitive identification, semantic-geometric alignment, and dynamic self-verification.

\noindent\textbf{Object-Centric Semantic Primitive Identification} 
We initiate with task scenario construction and subgoal decomposition to identify semantically critical primitives in an object \( o \). Specifically, we use VLMs to analyze geometric and physical features from multi-view images (\( \mathcal{I} \)) to infer potential manipulation tasks (\( \mathcal{T} = \{t_1,...,t_m\} \)). Each task \( t_i \) is decomposed into sub-stages with explicit operation goals per stage:
\begin{equation}
    \mathcal{G}_i = \{g_{i1},...,g_{ik}\}
    \label{eq:subgoals}
\end{equation}
To identify task-relevant spatial primitives, we establish primitive constraints for each subgoal:
\begin{equation}
    \mathcal{R}_{ij}^o(P_{ij}^o, A_{ij}^o) \implies g_{ij}
    \label{eq:constraint}
\end{equation}
where \( P_{ij}^o \subseteq \mathcal{P}^o \) denotes point primitives and \( A_{ij}^o \subseteq \mathcal{A}^o \) axis primitives for object \( o \). The unified primitive set is obtained through:
\begin{equation}
    \mathcal{E}^{o} = \left( \bigcup_{i,j} P_{ij}^o,\; \bigcup_{i,j} A_{ij}^o \right)
    \label{eq:unified_primitives}
\end{equation}
Notably, natural language descriptions of constraints (\( \mathcal{R}_{ij}^o \)) and primitives (\( \mathcal{E}^{o} \)) enable parallel semantic-geometric processing.

\noindent\textbf{Visual-Semantic Primitive Alignment} To align task-driven geometric primitives with visual annotated primitives from Sec~\ref{sec:geom_primitive_extraction}, we leverage VLM's multimodal geometric reasoning for mapping. For semantic keypoints,  we employ a multi-candidate matching strategy: when multiple geometric keypoints (\( p_i \in \mathcal{K}^{o}_{\text{filter}} \)) match semantic descriptions in \( \mathcal{E}^{o} \) in object \( o \) (e.g., "handle center" may correspond to multiple points detected on the handle), we record all candidates with confidence scores(\(s \in [0,1] \)) to ensure operational  robustness and handle occlusions. Unmatched cases return \textit{NONE}. For semantic orientations, we adopt dual representation: 1) symbolic axis descriptions (e.g., +Z-axis for vertical orientation), 2) flexible point-pair directions (e.g., spout direction from spout base to tip). Through cross-view consistency, we ensure point label persistence across perspectives, generating robust spatial-semantic mappings with 2D/3D annotations as shown in Fig.~\ref{fig:overview}.

\begin{algorithm}
\caption{Dynamic Self-Refine Matching Mechanism} 
\label{alg:dynamic_refinement}
\textbf{Input}: 
\begin{itemize}
    \item Object $\mathcal{O}_i$ 
    \item Segmentation granularity levels $\Gamma = \{\gamma_1,\dots,\gamma_N\}$ 
    \item Geometric primitives $\mathcal{K}_i = \{\mathcal{P}_i^{(1)}, \mathcal{A}_i^{(1)}, \dots\}$ 
    \item Semantic primitives $\mathcal{E}_i = \{\mathcal{P}_i^{sem}, \mathcal{A}_i^{sem}\}$ 
    \item Correspondence set $\mathcal{C}_i = (\mathcal{E}_i, \mathcal{K}_i^{(N)}, {s}_i)$
\end{itemize}
\textbf{Output}: Refined correspondence set $\hat{\mathcal{C}}_i$ or Matching Failure

\begin{algorithmic}[1]
\State Initialize $\gamma \gets \gamma_1$, $t \gets 0$, $\tau_{max} \gets 5$
\While{$t < \tau_{max}$}
    \State $t \gets t + 1$
    \State \textbf{Match:} Evaluate confidence scores ${s}_i$ in correspondence set $\mathcal{C}_i$
    \State $\mathcal{L} \gets \{ e_j \in \mathcal{E}_i \mid s_j < 0.5 \lor \text{unmatched} \}$
    \If{$\mathcal{L} = \emptyset$}
        \State \Return $\hat{\mathcal{C}}_i = (\mathcal{E}_i, \mathcal{K}_i^{(\gamma)}, {s})$
    \Else
        \If{$t = \tau_{max}$ \textbf{or} $\gamma = \gamma_{\text{max}}$}
            \State \Return Matching Failure
        \EndIf
        \State $\gamma \gets \gamma^+$ \Comment{Update to the next (finer) granularity level}
        \State \textbf{Resample:} $\mathcal{K}_i^{(\gamma)} \gets \text{SemanticSAM}(\mathcal{O}_i, \gamma)$
        \State \textbf{Align:} $\mathcal{C}_i^{(\gamma)} \gets \text{VLM}(\mathcal{L}, \mathcal{K}_i^{(\gamma)})$
        \State \textbf{Refine:} $\hat{\mathcal{C}}_i \gets \text{ReplaceLowConf}(\mathcal{C}_i^{(\gamma)}, \mathcal{L})$
    \EndIf
\EndWhile
\State \Return Matching Failure
\end{algorithmic}
\end{algorithm}

\begin{figure*}[t]
    \centering
    \vspace{-10pt}
    \includegraphics[width=0.98\linewidth]{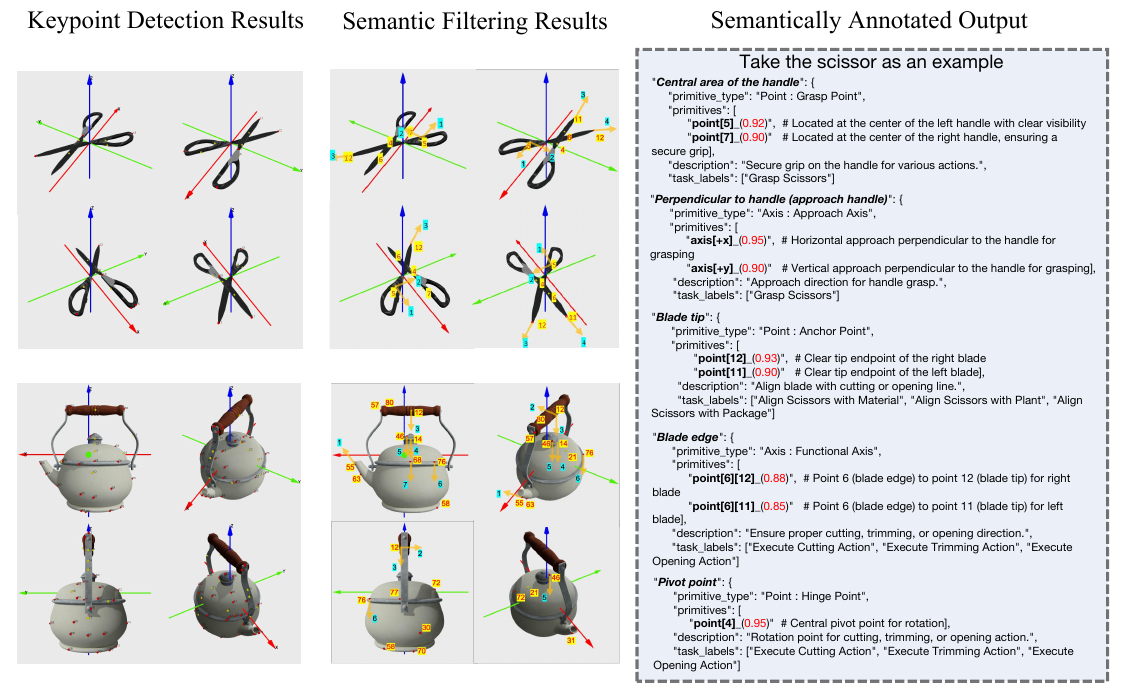} 
    \vspace{-10pt}
    \caption{Keypoints and Axes Annotated Output. This framework demonstrates the process of detecting, filtering, and semantically annotating functional keypoints and axes on 3D objects. The visualization progresses from initial keypoint detection (left column) to semantic filtering (middle column), culminating in rich semantic annotation (right column).}
    \label{fig:annotated output}
    \vspace{-6pt}
\end{figure*}

\noindent\textbf{Dynamic Self-Refine Matching Mechanism}
To address the inaccuracies in primitives detection and information loss caused by feature extractors, we adopt a dynamic self-refine matching algorithm, as shown in Algorithm~\ref{alg:dynamic_refinement}. First, lexical pattern matching validates annotation mappings. When low-confidence predictions ($<0.5$) or missing primitives (labeled as \textit{NONE}) are detected, the framework triggers hierarchical segmentation-based incremental annotation. Leveraging Semantic SAM's multi-granularity segmentation, this process refines geometric primitives via adaptive resampling, forming a closed-loop dynamic optimization workflow: \textit{segment-align-detect-resample}. Experiments demonstrate that our method achieves a 98\% matching success rate on our dataset and effectively mitigates error propagation from poor segmentation.

\subsection{Semantic-guide Reasoning in Manipulation}
Beyond generating geometrically annotated object datasets, our framework facilitates the integration of spatial semantics into manipulation tasks. By leveraging multi-modal inputs—including geometrically annotated objects and their corresponding hierarchical semantics—PASG provides semantic-aware visual cues and textual guidance to downstream reasoning modules.

Specifically, through our manipulation experiments (in Section~\ref{sec:manipulation_experiment}), we validate that the PASG pipeline can reliably identify interaction primitives across diverse object categories. Compared to manually labeled counterparts, PASG-generated annotations exhibit greater semantic diversity (e.g., multiple meaningful grasp or actuation points). This diversity contributes to flexible downstream usage and competitive task success rates across varied manipulation tasks, demonstrating the practical effectiveness of PASG in generating semantically grounded geometric annotations for real-world deployment.

\section{Experiment}
\label{sec:experiment}

\subsection{Semantic-aware Object Dataset}

\noindent\textbf{Data Sources and Scales}
We collect raw datasets sourced from RoboCasa \citep{nasiriany2024robocasalargescalesimulationeveryday} and Objaverse \citep{Deitke_2023_CVPR, NEURIPS2023_70364304}. RoboCasa provides over 2,500 high-quality 3D objects covering more than 150 categories in everyday tasks, whereas Objaverse is a large-scale open dataset containing over 800,000 annotated 3D objects. Since our focus is on objects with meaningful spatial semantics, we filtered the 3D assets based on their tags and titles, excluding objects with narrowly defined functionalities or limited manipulability potential (e.g., food items). Leveraging texture detection, we further refined our selection to obtain a high-quality dataset of 5,231 objects.
To ensure effective object segmentation, all selected objects are rescaled to appropriate sizes and rendered from multiple viewpoints, including four horizontal orthographic views and four 45-degree oblique orthographic projections. In total, we acquired a 5,231 object 3D dataset as well as an 41,848 image 2D dataset for subsequent semantic annotation.

\noindent\textbf{Data Annotation}
As mentioned in Section~\ref{sec:geom_primitive_extraction}, we first utilize VFM combined with topological methods to extract and visualize key primitives from multi-view object images. To ensure consistent and effective visual labeling, we perform the projection of 2D keypoints onto their corresponding 3D objects and the subsequent disambiguation and filtering through Open3D \citep{zhou2018open3dmodernlibrary3d}. Each annotated point is assigned a visual index to facilitate identification. Leveraging the strong semantic reasoning capabilities of VLM, we feed the annotated multi-view images into GPT-4o \citep{openai2024gpt4ocard} as a visual prompt for spatial semantic annotating, as detialed in Section~\ref{sec:task_semantic_annotation}. Based on annotations, we filter out primitives that lack critical semantic information, producing the final set of semantic annotations for each object.

\noindent\textbf{Dataset Validation}
To evaluate the effectiveness of our dataset generation pipeline, we randomly sample 50 annotated objects and conduct a manual inspection. Our experiment shows that, during the semantic identification phase, GPT-4o achieves an accuracy of 91.6\%. During the geometric-semantic alignment phase, it demonstrates strict alignment accuracy of 75.8\% (requiring full primitive correctness) and alignment effectiveness of 91.5\% (ensuring at least one valid primitive-semantic match). These findings highlight GPT-4o's strong spatial reasoning abilities under structured prompts, confirming the reliability of our annotated dataset. However, complex spatial semantics of the objects remain a challenge for GPT-4o. To further improve annotation quality, we employ a multi-round validation strategy where GPT-4o iteratively self-corrects its annotations, enhancing the robustness of the dataset’s semantic labels. Additional validation experiments are provided in the supplemental material.

\subsection{Manipulation Task Evaluation}
\label{sec:manipulation_experiment}
To validate the effectiveness of PASG in robotic manipulation, we conduct comprehensive evaluations using the RoboTwin\citep{mu2024robotwindualarmrobotbenchmark} simulation platform, open-source environment designed to emulate realistic robotic manipulation scenarios. RoboTwin provides standardized benchmarks that ensure both reproducibility and practical relevance.

\noindent\textbf{Task Setting}
We evaluate PASG's performance across six representative manipulation tasks, including tasks requiring dual-arm coordination and single-arm manipulations, as well as interactions within cluttered environments. Specifically, the tasks are: (1) Block Hammer Beat, (2) Container Place, (3) Dual Bottles Pick, (4) Empty Cup Place, (5) Pick Apple, and (6) Messy Shoe Place. Detailed descriptions of each setup are provided in the supplemental material.

\noindent\textbf{Quantitative Results}
We quantitatively compare the task success rates of PASG against a baseline involving manual annotations. While human annotations are manually curated and serve as the gold standard, PASG annotations are generated automatically with minimal manual filtering. Each task is executed 100 times using randomly initialized seeds to ensure robustness of the evaluation. Results of this comparison are summarized in Table~\ref{tab:manipulation_performance}, the PASG-based policy achieves competitive performance compared to manual annotations, and even outperforms them in tasks such as “Block Hammer Beat” and “Empty Cup Place”. These results demonstrate the ability of our pipeline to generate functional and accurate geometric primitives for downstream manipulation.

\begin{table}[ht]
\centering
\scriptsize
\tabcolsep3pt
\resizebox{0.95\linewidth}{!}{
\begin{tabular}{c | c c c c c c | c}
\toprule
\textbf{Method} & \textbf{BHB} & \textbf{CP} & \textbf{DBP} & \textbf{ECP} & \textbf{PAM} & \textbf{SP} & \textbf{Avg.} \\
\midrule
Human Annotation & 79.0 & 93.0 & 95.0 & 73.0 & 85.0 & 83.0 & 84.67 \\
PASG & \textbf{82.0} & 89.0 & 70.0 & \textbf{76.0} & 81.0 & 69.0 & 77.83 \\
\bottomrule
\end{tabular}}
\vspace{-6pt}
\caption{
\footnotesize{Task success rates (\%) for different manipulation scenarios. Bold highlights where PASG outperforms human annotations.}}
\label{tab:manipulation_performance}
\vspace{-10pt}
\end{table}

\begin{figure}[h]
\vspace{-8pt}
\centering 
\includegraphics[width=0.98\linewidth]{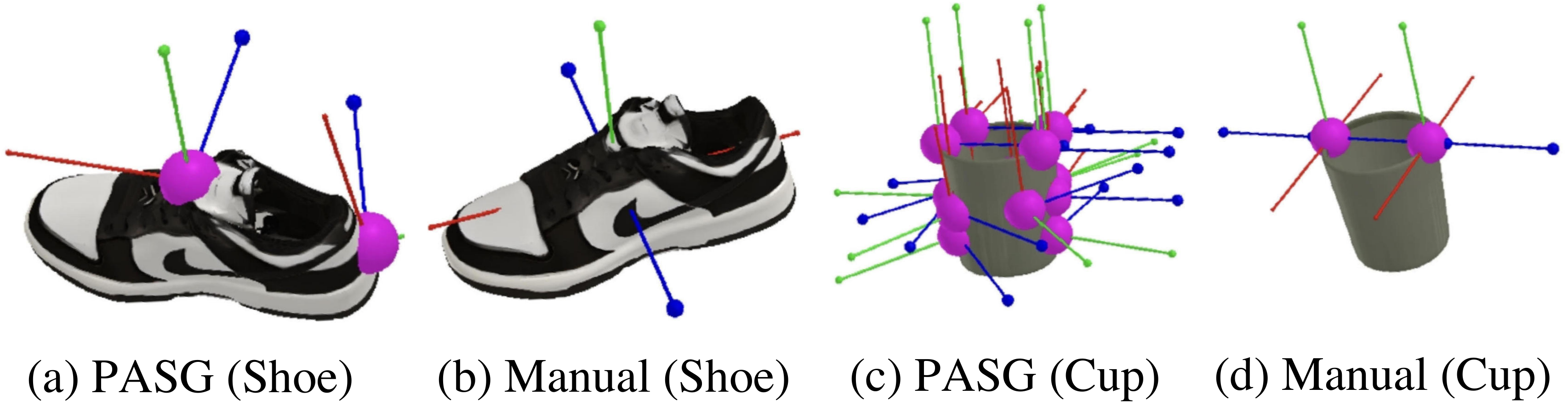}
\vspace{-6pt}
\caption{Compared to manual annotation, PASG tends to generate a more diverse and semantically accurate set of interaction points.}
\vspace{-15pt}
\label{fig:vis-comparison} 
\end{figure}

\noindent\textbf{Qualitative Results}
A key advantage of PASG is its ability to generate a richer and more diverse set of interaction primitives compared to manual annotation. Human labelers, constrained by cost and effort, tend to identify only a few optimal points. In contrast, our automated framework identifies a wider array of semantically meaningful points, as illustrated in Figure~\ref{fig:vis-comparison}. This diversity provides the manipulation policy with greater flexibility and enhances robustness to variations in task execution. For instance, the availability of multiple valid grasp points on a shoe or a cup allows successful execution even in cases of occlusion or restricted access.

\subsection{Object-based Spatial-Semantic Reasoning}

\begin{table*}[htbp]
  \small 
  \centering
  \begin{tabular}{@{}l|cccc|cccc@{}}
    \toprule
    \multicolumn{1}{c|}{} & \multicolumn{4}{c|}{in-distribution test set} & \multicolumn{4}{c}{out-of-distribution test set} \\
    \cmidrule(lr){2-5} \cmidrule(lr){6-9}
    Model & Task1 & Task2 & Task3 & Overall & Task1 & Task2 & Task3 & Overall \\
    \midrule
    GPT-4V \citep{openai2034gpt4vcard}     & 32.92 & 38.14 & 46.1 & 39.04 & 37.56 & 36.35 & 43.93 & 39.00 \\
    GPT-4O \citep{openai2024gpt4ocard}     & 37.87 & 40.07& 52.48 & 43.19 & 38.29 & 42.99 & 43.69 & 41.79 \\
    GPT-4O-mini \citep{openai2024gpt4ocard} & 37.32 & 30.26 & 29.85 & 32.26 & 37.87 & 27.59 & 42.08 & 34.96 \\
    LLaVA-1.5 \citep{liu2023improvedllava}  & 30.20 & 33.04 & 32.15 & 31.95 & 31.95 & 25.09 & 36.89 & 30.72 \\
    Claude-3.5 \citep{anthropic2024claude3.5} & 29.95 & 35.68 & 34.28 & 33.6 & 25.12 & 34.13 & 36.17 & 32.04 \\
    Qwen-2.5VL \citep{bai2025qwen25vltechnicalreport} & 50.00 & 36.56 & 47.99 & 43.91 & 46.10 & 43.73 & 40.05 & 43.33 \\
    \midrule
    SpaceMantis \citep{jiang2024mantis} & 37.13 & 10.54 & 46.57 & 29.15 & 33.66 & 8.86 & 26.70 & 21.70 \\
    RoboPoint \citep{yuan2024robopoint}  & 32.67 & 0 & 16.31 & 14.40 & 34.63 & 0 & 16.26 & 15.32 \\
    \midrule
    Qwen2.5VL-PA & \textbf{86.63} & \textbf{83.48} & \textbf{61.70} & \textbf{77.79} & \textbf{89.02} & \textbf{81.73} & \textbf{67.72} & \textbf{79.69} \\
    \bottomrule
  \end{tabular}
  \caption{Spatial comprehension evaluation on our visual question-answer benchmark. Numbers represent accuracy (\%).}
  \label{tab:fine_tune}
  \vspace{-12pt}
\end{table*}

\noindent\textbf{Benchmark}
To evaluate whether our framework effectively captures spatial primitives (keypoints and axes), we developed Robocasa-PA, a visual question-answer benchmark derived from the Robocasa dataset using PASG. This spatial-aware benchmark evaluates models’ understanding of functional geometric primitives in robotic manipulation scenarios. The dataset comprises three question categories: Task 1 \textit{Type Identification} (determining the functional category of spatial primitives from visual features), Task 2 \textit{Task Association} (linking detected primitives to specific manipulation tasks), and Task 3 \textit{Task-to-Primitive Mapping} (identifying primitives required to accomplish given tasks). These three question types collectively assess both geometric perception (e.g., structural recognition of primitives) and task-aware reasoning (e.g., understanding the functional significance of primitives in context).

The dataset is divided into three parts for evaluating model generalization. We first generate 6,979 questions from a designated pool of base objects, allocating 80\% (5,583 questions) as the fine-tuning training set to establish a foundational understanding of primitive structures. The remaining 20\% (1,396 questions) of the same object pool formed the in-distribution test set. To rigorously assess cross-domain adaptability, we introduce an out-of-distribution test set comprising 1,364 questions exclusively derived from unseen objects, ensuring strict isolation from training instances at both object and primitive levels. All images are validated to ensure visibility of the referenced primitives. Each question follows a single-choice format, and accuracy is used to evaluate performance across all sets.

\noindent\textbf{Finetune}
We fine-tuned Qwen-2.5VL \citep{bai2025qwen25vltechnicalreport} using Low-Rank Adaptation (LoRA) to assess whether the VQA benchmark supports knowledge transfer in primitive compositional reasoning. By constraining parameter updates to low-rank decomposition matrices, performance improvements can be causally attributed to knowledge distillation from the benchmark. We selected several VLMs as baselines, including general-purpose large-scale vision language models and models with spatial awareness capabilities proposed in prior works, as shown in Table \ref{tab:fine_tune}.

The fine-tuned Qwen2.5VL-PA shows significant improvements over the baselines, validating our framework’s ability to distill geometric-semantic knowledge and task-aware reasoning. Notably, the model exhibits consistent robustness in out-of-distribution tests, showcasing enhanced cross-domain adaptability and task-transfer potential.

\begin{figure}[htbp]
    \vspace{-5pt}
    \centering
    \includegraphics[width=\columnwidth]{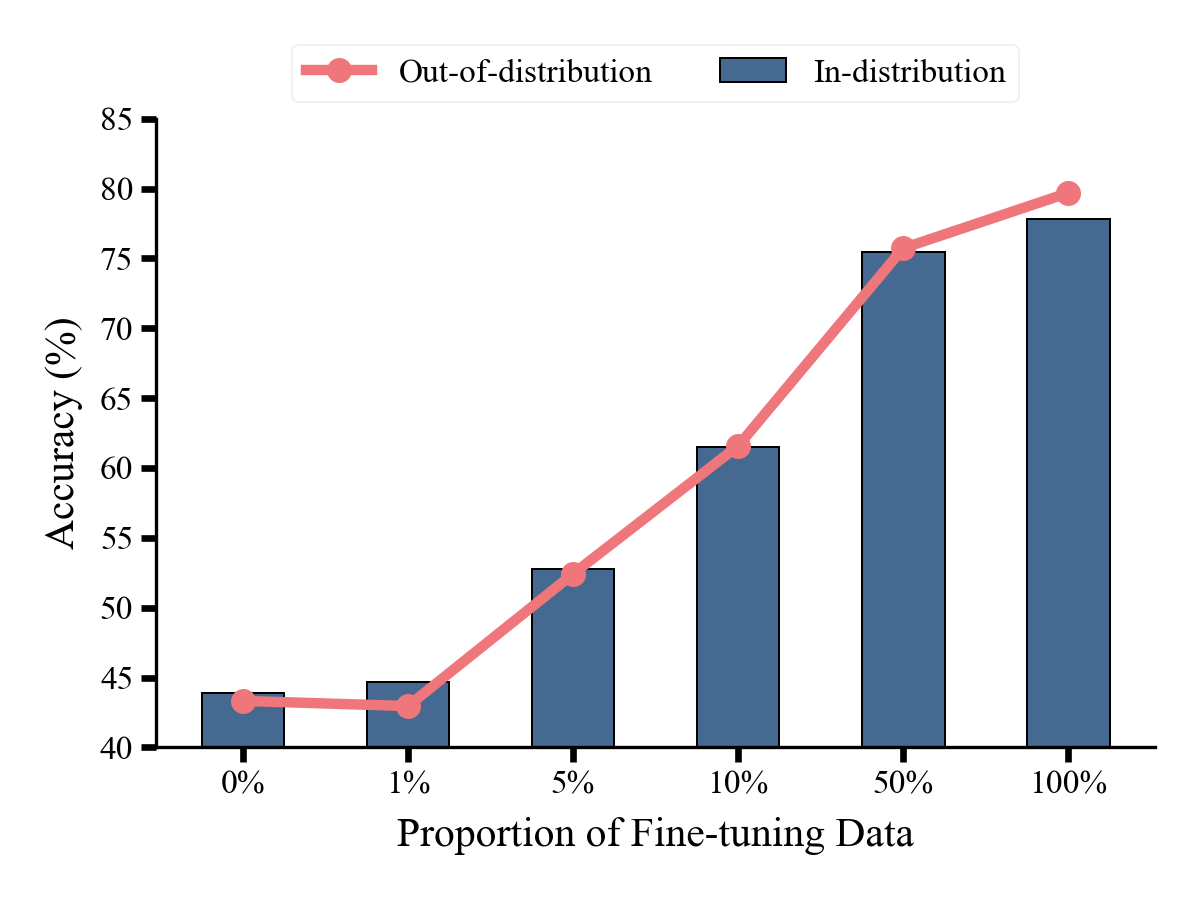} 
    \vspace{-18pt}
    \caption{Data Effectiveness Study} 
    \label{fig:data_effectiveness_study}
    \vspace{-6pt}
\end{figure}

\noindent\textbf{Data Effectiveness}
To evaluate the effectiveness of fine-tuning data, we conducted a progressive scaling experiment: fine-tune the model with randomly sampled subsets of 1\% (55 samples), 5\% (279 samples), 10\% (558 samples), and 50\% (2,791 samples) from the original training set. As shown in Fig \ref{fig:data_effectiveness_study}, with only 5\% data, the model achieved an absolute accuracy improvement of approximately 10\% on both in-distribution and out-of-distribution test sets, corresponding to a 20.6\% relative improvement over the original non-fine-tuned model. When increasing the subset to 10\%, the absolute accuracy further improved by 20\%, achieving a 41.12\% relative improvement over the baseline. Notably, the performance gap between out-of-distribution and in-distribution test sets remained stable within ±2\%. These results demonstrate that our proposed data filtering pipeline can efficiently extract high-value data, significantly reducing annotation costs while ensuring robust cross-distribution generalization.

\section{Conclusion}
\label{sec:conclusionandfuturework}
This paper presents PASG, a closed-loop framework that bridges task semantics and geometric primitives in robotic manipulation. By combining automated geometric primitive extraction with VLM-driven semantic anchoring, PASG enables spatial-semantic reasoning in unstructured environments. It overcomes key limitations in existing systems through geometry-aware feature aggregation, dynamic coupling of primitives with functional affordances, and self-corrective mechanisms to reduce error propagation. Evaluations on the RoboTwin platform across diverse manipulation tasks show PASG performs competitively with human annotations, even outperforming them in certain tasks. PASG’s ability to generate diverse interaction primitives enhances task flexibility and robustness, making it suitable for real-world applications. Additionally, the Robocasa-PA benchmark and fine-tuned Qwen2.5VL-PA model demonstrate the framework’s effectiveness, providing an automated pipeline for generating high-quality annotated data that improves generalization and cross-domain adaptability in robotic manipulation.

\clearpage
\noindent\textbf{Acknowledgement} This work was supported by the Shanghai Municipal Science and Technology Major Project (2021SHZDZX0102), and the Fundamental Research Funds for the Central Universities.

{
    \small
    \bibliographystyle{ieeenat_fullname}
    \bibliography{main}
}

\newpage
\appendix
\section*{Supplemental Material}
\section{Prompt Design}
\label{sec:supp_hardware}
\subsection{Annotate Prompt}
\begin{tcolorbox} [colback=gray!10, colframe=black, arc=0mm, width=\textwidth, enlarge left by=-10mm,]
Control a robot to perform manipulation tasks based on an image of a single object with marked keypoints and a text instruction. The goal is to list the possible uses of the object and all key primitives in each stage.
    
    Object Analysis
    
    - Determine how many important parts of the object based on the usage and image. For example:
    \quad - "Important parts of the Teapot":
    
        \qquad - Body: The main container for holding the tea or water.
        
        \qquad- Lid: A cover for the opening at the top of the teapot to prevent spillage and retain heat.
        
        \qquad- Handle: A grip for holding and pouring the teapot, often designed to insulate from heat.
        
        \qquad- Spout: A narrow outlet for pouring the liquid from the teapot.
        
        \qquad- Base: The bottom support of the teapot that ensures stability and contact with the surface. \\

    Possible Uses
    
    - According to the important parts above, determine how many usages of the object in image. For example:
    
      \quad - "Possible Uses of the Teapot":
      
        \qquad- Pouring tea from the teapot.
        
        \qquad- Filling the teapot with water.
        
        \qquad- Place the fallen teapot upright.\\

    Task Stages
    
    - Break down each use into stages. For example:
    
      \quad - "pouring tea from teapot":
      
        \qquad - 3 stages: "grasp teapot", "align teapot with cup opening", and "pour liquid"
        
       \quad - "put red block on top of blue block":
       
        \qquad - 2 stages: "grasp red block", "drop the red block on top of blue block"
      
      \quad - "reorient bouquet and drop it upright into vase":
      
        \qquad - 3 stages: "grasp bouquet", "upright bouquet", and "align the reoriented bouquet with vase"
        
      \quad - "open microwave door to heat food":
      
        \qquad - 6 stages: "grasp microwave door handle", "pull door open along hinge axis", "grasp food", 
        
        \qquad "place food container on turntable", "close door along hinge axis" and "press start button"\\
    
    Key Primitive Definitions 
    
    - List all candidate key primitives involved in each stage categorized as:
    
        \qquad - *Main*: The reference point and principal orientation axis of the object, serving as the basis for defining its spatial configuration. (e.g. Center of the teapot body + vertical axis.)
        
        \qquad - *Anchor*: A specific point and axis used as a reference for defining the object's pose during movement or ensuring keypoint constraints are met at the end of a substage (e.g., Tip of the spout + pouring direction.).
        
        \qquad - *Grasp*: The position and orientation of the end-effector when securely holding the object (e.g., Center of the handle + approach direction for grasping).
        
        \qquad - *Actuation*: The position and orientation of the end-effector required to trigger mechanical operations, such as pressing, rotating, or toggling components (e.g., Center of the heating button + pressing direction).
        
        \qquad- *Hinge*: The position and orientation of the end-effector used to manipulate articulated objects, typically around a rotational axis (e.g., Center of the lid hinge + rotation axis direction).\\        

\end{tcolorbox}

\clearpage

\begin{tcolorbox} [colback=gray!10, colframe=black, arc=0mm, width=\textwidth, enlarge left by=-10mm,]
        \qquad - *Actuation*: The position and orientation of the end-effector required to trigger mechanical operations, such as pressing, rotating, or toggling components (e.g., Center of the heating button + pressing direction).
        
        \qquad- *Hinge*: The position and orientation of the end-effector used to manipulate articulated objects, typically around a rotational axis (e.g., Center of the lid hinge + rotation axis direction).\\
      
    Key Primitive Example
    
    - Define key primitives for each stage. For example:
    
        key\_primitives = [
            
            \{Type: Main, Pos: [x, y, z], Orientation: [dx, dy, dz], Stage: "Pour Liquid", Description: "Global reference for maintaining proper pouring orientation"\},
            
            \{Type: Grasp, Pos: [x, y, z], Orientation: [dx, dy, dz], Stage: "Grasp Teapot", Description: "Grasping the teapot handle for secure hold"\},

            \{Type: Anchor, Pos: [x, y, z], Orientation: [dx, dy, dz], Stage: "Align Teapot with Cup Opening", Description: "Reference point and axis for positioning the teapot relative to the cup"\}
            
            \# \quad Add more key primitive if needed
        ]\\

    **Note:**
    
    - You do not need to consider collision avoidance. Focus on what is necessary to complete the task.
    
    - *List all possible options thoroughly* when there are multiple reasonable candidate primitives.
    
    - If the object has multiple possible uses, try to cover as many usage scenarios as possible, rather than just the most obvious operation.
    
    - When you annotate a point, you MUST label its corresponding orientation; vice versa. 
    
    - Prioritize vectors based on the object's skeleton or intrinsic orientation, and then consider vectors representing orientations that are not directly visible in the image.
    
    - Do not contain the specific id of key points in the your output.
    
    - The initial orientation may affect grasping points, annotate all available primitives. For example, if handles are unusable, consider edges for grasping.
    
    - Do not use '...' for omission in your output, list all usages and possibles.
    
    - If the task involves multiple sub-operations, ensure that all stages are clearly labeled and avoid omitting any possible intermediate steps (such as rotation, alignment, etc.)
    
    - List all possible uses of the object, ensuring diversity and avoiding repetitive scenarios by considering:
    
        \qquad - Interaction with typical objects (how the object interacts with other common objects in manipulation).
        
        \qquad - Position or orientation adjustment (how the object's position or orientation is adjusted for a specific purpose).
        
        \qquad - Functional Demonstration (describe tasks that showcase the object's functionality, such as pressing buttons, turning knobs, or activating features to achieve a specific outcome.).
        
    - Pay attention to identifying the **Actuation Point** of any special buttons or knobs. For example, use the button to open the lid before heating a liquid (or filling with water). 
    
        \qquad - There are two sub-steps(locate the button then press the button) and two key primitives(button center area and press orientation).
        
        \qquad- If there are multiple buttons, please make sure to distinguish between them.(eg. the heating button or the lid opening button)
        
    - Special consideration for Actuation (execution) points: Differentiate between various types of control objects such as buttons, knobs, sliders, and other mechanical interfaces.
    
    - In this stage, you don't need to fill in the specific value in Pos and Orientation, just let it be [x, y, z] and [dx, dy, dz] \\
        
    **Structure of python code block as follows:**
    
    \quad List all important parts of the object based on the usage and image
    
\end{tcolorbox}

\clearpage

\subsection{Alignment Prompt}
\begin{tcolorbox} [colback=gray!10, colframe=black, arc=0mm, width=\textwidth, enlarge left by=-10mm,]
    You are provided with semantic annotations of an object in the image and need to map these annotations to the corresponding primitives detected in the image. Your goal is to generate a JSON annotation result that aligns the semantic labels with the detected keypoints based on the following rules: \\
   
    The Definition of Principal Orientation
   
    - The principal orientation (X, Y, Z) are marked in different colors in the image. Ensure that you match the following colors to the corresponding axes:
    
        \qquad - X Axis: Red, +x: [1, 0, 0]; -x: [-1, 0, 0]
    
        \qquad - Y Axis: Green, +y: [0, 1, 0]; -y: [0, -1, 0]
     
        \qquad  - Z Axis: Blue, +z: [0, 0, 1]; -z: [0, 0, -1]
    \\
    
      Key Primitive Mapping
      
      - If a key position is correctly identified, include the corresponding point in the "points" attribute. 
      
      - Append a probability value to each point in the format 
      \{"pos\_ID": n, "pos\_Probability": p\}. 
      
      For example: 
      
      \qquad \{"pos\_ID": 1, \quad "pos\_Probability": 0.75\}, 
      
      \qquad \{"pos\_ID": 2, \quad "pos\_Probability": 0.85\}. 
      
      - If multiple points represent the same position, list them all.
      
      - If a key orientation is correctly identified, represent it with following two ways: 
      
        \qquad - Using Principal Orientation Labels for standard orientations (e.g., horizontal/vertical):  
        
        \qquad   [0, 0, 1], [1, 0, 0]. For example, [0, 0, 1] means orientation towards the positive z-axis.
        
        \qquad - Using Point-based orientation for object-specific orientations (e.g., spout orientation): 
        
        \qquad   [x, y] (x, y is the ID of the keypoints, and the orientation is from point X to point)
        
      - Append a probability value to the orientation in the format \{"ori\_ID": [x, y], "ori\_Probability": p\} or \{"ori\_ID": [0, 0, 1], "ori\_Probability": p\}. For example: \{"ori\_ID": [1, 2], "ori\_Probability": 0.7\}, \{"ori\_ID": [0, 0, 1], "ori\_Probability": 1.0\}. If there are multiple pairs for the same orientation, list them all. \\
      
    - Probability values indicate the confidence level of the point/orientation being correctly identified:
    
      \qquad $>$ 0.5: The point/orientation is worth considering but may need verification.
      
      \qquad $>$ 0.8: The point/orientation is highly accurate and can be trusted.
      
    - Handling Missing or Erroneous Annotations:
    
        \qquad - Use "None" if the keypoint/orientation is not visible in the viewpoints.
      
        \qquad - Use "Error" if the keypoint/orientation is visible in the viewpoints, but no points annotated. (Re-detect is required) \\

    **Note:**
    
    - Exclude Undetected Points: Do not include points that are not present in the image.
    
    - Output Format: Only output the final JSON result without any additional text.
    
    - The initial orientation may affect grasping points, annotate all available positions. For example, 
    if handles are unusable, consider edges for grasping.
    
    - Identify and list as many effective points as possible.
    
    - The point numbers are continuous and consistent across different viewpoints. Match the correct points and vectors by aligning the point numbers across different perspectives.
    
    - "ori\_ID": [x, y] (towards the target) and "ori\_ID": [y, x] (away from the target).
    
    - Approach Orientation represents the end-effector movement towards the Grasp Point or Actuation Point. Ensure orientation (e.g., "ori\_ID": [x, y] or "ori\_ID": [0, 0, -1]) point towards the target, not away from it.
      - For example, USE "ori\_ID": [x, y] (towards the handle/neck), NOT "ori\_ID": [y, x] (away from the handle/neck), to avoid moving the end-effector away from the manipulation point.
      
    - In this stage, you don't need to fill in the specific value in Pos and Orientation, just let it be [x, y, z] and [dx, dy, dz] \\
    
\end{tcolorbox}

\clearpage
\begin{tcolorbox} [colback=gray!10, colframe=black, arc=0mm, width=\textwidth, enlarge left by=-10mm,]
    
    if handles are unusable, consider edges for grasping.
    
    - Identify and list as many effective points as possible.
    
    - The point numbers are continuous and consistent across different viewpoints. Match the correct points and vectors by aligning the point numbers across different perspectives.
    
    - "ori\_ID": [x, y] (towards the target) and "ori\_ID": [y, x] (away from the target).
    
    - Approach Orientation represents the end-effector movement towards the Grasp Point or Actuation Point. Ensure orientation (e.g., "ori\_ID": [x, y] or "ori\_ID": [0, 0, -1]) point towards the target, not away from it.
      - For example, USE "ori\_ID": [x, y] (towards the handle/neck), NOT "ori\_ID": [y, x] (away from the handle/neck), to avoid moving the end-effector away from the manipulation point.
      
    - In this stage, you don't need to fill in the specific value in Pos and Orientation, just let it be [x, y, z] and [dx, dy, dz] \\

    Output Format
    
    - Organize the annotations by type and format them in the required JSON structure:\\
    **JSON format your output for task description and key primitives**
    
    \{
    
        \qquad "Grasp": [
        
            \qquad \qquad \{
            
                \qquad  \qquad \qquad "Stage": "Grasp Teapot",
                
                \qquad  \qquad \qquad "pos\_ID": n,
                
                \qquad  \qquad \qquad "pos\_Probability": p\_1,
                
                \qquad  \qquad \qquad "ori\_ID": [x, y],
                
                \qquad  \qquad \qquad "ori\_Probability": p\_2
                
                \qquad  \qquad \qquad "Pos": [x, y, z],
                
                \qquad  \qquad \qquad "Orientation": [dx, dy, dz],
                
                \qquad  \qquad \qquad "Description": "Grasping the teapot handle for secure hold"
                
            \qquad \qquad \},

        \qquad],
        
        \qquad "Anchor": [
        
           \qquad \qquad \{
           
                \qquad  \qquad \qquad"Stage": "Align Teapot with Cup Opening",
                
                \qquad  \qquad \qquad"pos\_ID": n,
                
                \qquad  \qquad \qquad"pos\_Probability": p\_1,
                
                \qquad  \qquad \qquad"ori\_ID": [0, 0, 1],
                
                \qquad  \qquad \qquad"ori\_Probability": p\_2,
                
                \qquad  \qquad \qquad"Pos": [x, y, z],
                
                \qquad  \qquad \qquad"Orientation": [dx, dy, dz],
                
                \qquad  \qquad \qquad"Description": "Reference point and axis for positioning the teapot relative to cup"
                
            \qquad \qquad \},
            
        \qquad ]
            
        \qquad "Hinge": [
        
            \qquad \qquad \{
            
                \qquad  \qquad \qquad "Stage": "Open Lid",
                
                \qquad  \qquad \qquad "pos\_ID": n,
                
                \qquad  \qquad \qquad "pos\_Probability": p\_1,
                
                \qquad  \qquad \qquad "ori\_ID": [0, 0, 1],
                
                \qquad  \qquad \qquad "ori\_Probability": p\_2
                
                \qquad  \qquad \qquad "Pos": [x, y, z],
                
                \qquad  \qquad \qquad "Orientation": [dx, dy, dz],
                
                \qquad  \qquad \qquad "Description": "Rotation center and axis for opening the lid"
                
            \qquad \qquad \}
            
        \qquad  ]
        
    \}
\end{tcolorbox}

\clearpage

\section{Implementation Details}
\subsection{Manipulation Task Evaluation}
\label{sec:manipulation_experiments}
To validate the effectiveness of PASG in robotic manipulation, we conduct comprehensive evaluations using the RoboTwin\citep{mu2024robotwindualarmrobotbenchmark} simulation platform, open-source environment designed to emulate realistic robotic manipulation scenarios. RoboTwin provides standardized benchmarks that ensure both reproducibility and practical relevance.

\begin{figure*}[t]
    \centering
    \includegraphics[width=\linewidth]{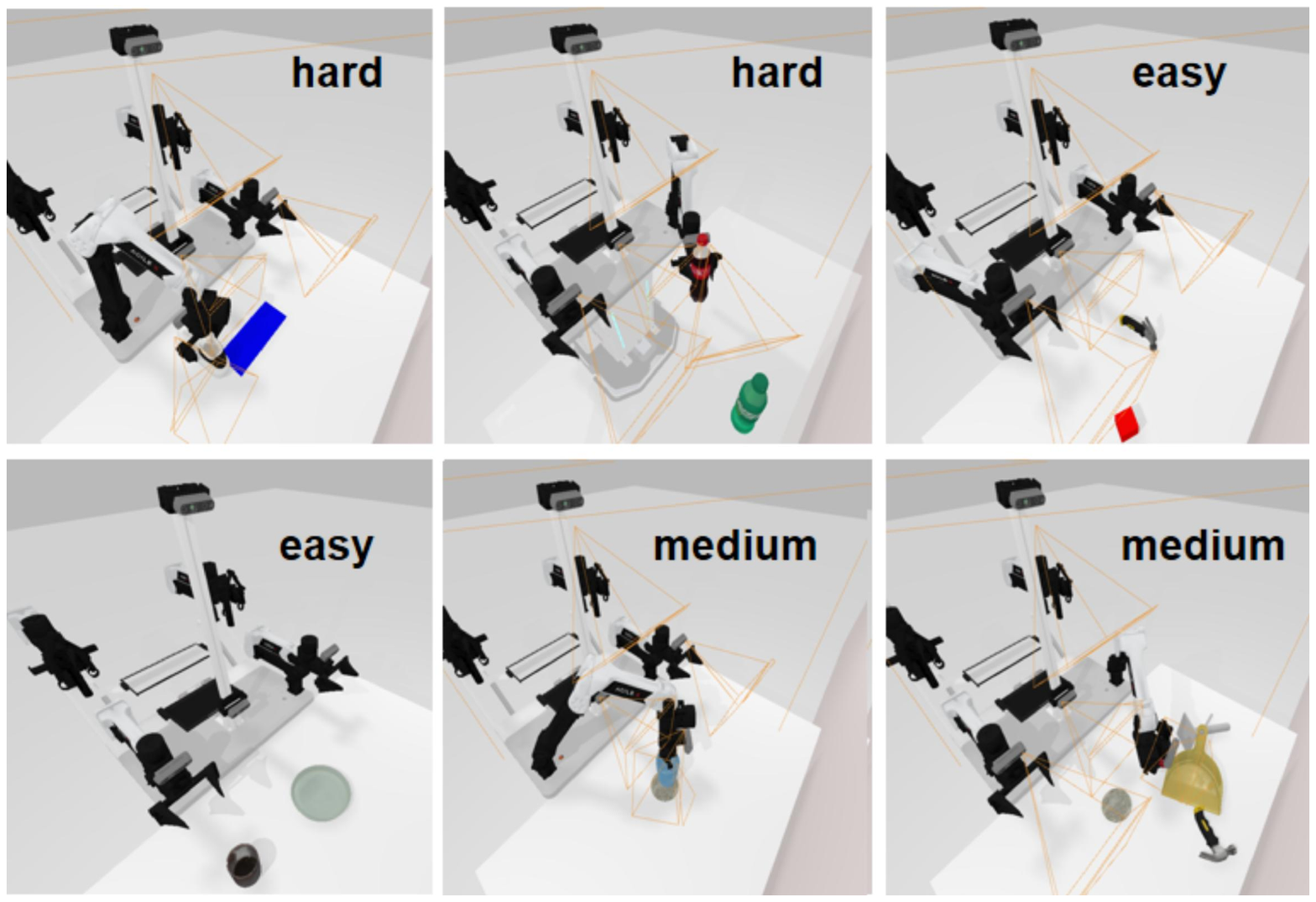} 
    \vspace{-20pt}
    \caption{Illustrations of the six representative manipulation tasks evaluated in PASG. From left to right and top to bottom, the tasks are: Messy Shoe Place, Dual Bottles Pick, Block Hammer Beat, Container Place, Empty Cup Place, and Pick Apple. The tasks are categorized by difficulty levels: the first two tasks are hard, the middle two are easy, and the final two are medium.} 
    \label{fig:experiment_overview}
    \vspace{-10pt}
\end{figure*}

\noindent\textbf{Task Setting}
We evaluate PASG's performance across six representative manipulation tasks, including tasks requiring dual-arm coordination and single-arm manipulations, as well as interactions within cluttered environments. Specifically, the tasks are: (1) Block Hammer Beat, (2) Container Place, (3) Dual Bottles Pick, (4) Empty Cup Place, (5) Pick Apple, and (6) Messy Shoe Place.(see Fig.~\ref{fig:experiment_overview} for visual illustrations and difficulty categories).

To provide a better understanding of these tasks, we briefly describe each one below, accompanied by an illustrative figure showcasing the robot's interactions. The tasks are categorized by difficulty levels: hard, medium, and easy, highlighting the range of challenges faced by PASG.

\begin{itemize}
    \item \textbf{Messy Shoe Place:} 
    The robot needs to grasp a shoe with one hand. The main difficulty lies in the fact that the shoe's width is nearly identical to the width of the gripper, resulting in a low tolerance for error.

    \item \textbf{Dual Bottles Pick:} 
    This task requires dual-arm coordination to simultaneously pick up two bottles and maintain balance while lifting them. It is a challenging task due to the necessity for precise synchronization between the two arms.

    \item \textbf{Block Hammer Beat:} 
    The robot uses a hammer to hit a block at a predefined position. This task evaluates fine motor skills and the ability to manipulate tools effectively.

    \item \textbf{Container Place:} 
    In this task, the robot picks up a container and places it accurately at a designated location. It is relatively easy because of the simple geometry of the object.

    \item \textbf{Empty Cup Place:} 
    The robot must handle a fragile and lightweight object (a cup) and place it gently. The challenge is to maintain stability while carefully avoiding excessive force.

    \item \textbf{Pick Apple:} 
    The robot is required to grasp a round and slippery object (an apple) from a cluttered environment. This is a medium-difficulty task that demands dexterity and precise perception.
\end{itemize}

\begin{figure*}[t]
    \centering
    \includegraphics[width=0.48\linewidth]{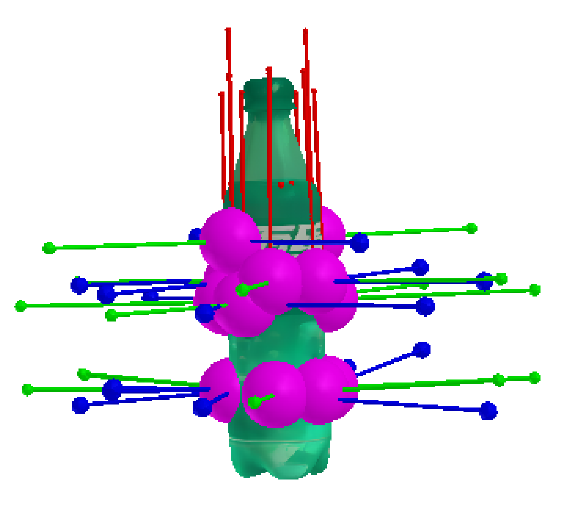}
    \hfill
    \includegraphics[width=0.48\linewidth]{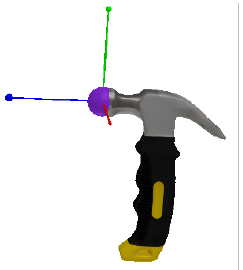}
    \caption{(Left) Automatically labeled grasping points on a bottle. The grasp points are distributed around the bottle's circular body, aligning with human intuition for stable and effective grasping. 
    (Right) Automatically labeled functional points on a hammer. The points indicate the striking area at the hammerhead, which corresponds to its intended use for hitting objects. These annotations demonstrate PASG's capability to identify functional and practical areas of tools autonomously.}
    \label{fig:tool_annotations}
\end{figure*}

\noindent\textbf{Automated Labeling Consistency with Human Usage} \\
The two images in Figure~\ref{fig:tool_annotations} illustrate PASG's ability to autonomously label tools in a way that aligns with human usage conventions. In the case of the bottle, the labeled grasping points are distributed evenly around its circular body, which ensures stability and effectiveness in gripping. For the hammer, the annotated functional points are concentrated on the hammerhead, which is the intended area for striking. These results highlight PASG's comprehensiveness and its capability to capture both the functional and practical aspects of tools, closely mirroring human understanding of their usage.

\section{Additional Experiments}
\subsection{Reliability of VLM-Based Semantic Alignment}

To assess the robustness of our VLM-based annotation pipeline, we conducted a detailed hallucination analysis over 844 annotations generated by GPT-4o across 100 diverse objects. We categorize potential annotation errors into four types: (1) invalid label indices, (2) semantically unreasonable suggestions, (3) missing functional labels, and (4) output format violations.

\begin{table}[H]
\centering
\renewcommand\arraystretch{1.1}
\tabcolsep3.5pt
\resizebox{1.0\linewidth}{!}{
\begin{tabular}{c | c | c | c | c | c}
\toprule
\textbf{Metric} & \textbf{Invalid Label Index} & \textbf{Unreasonable Interaction} & \textbf{Missing Functional Labels} & \textbf{Format Violation} & \textbf{Total} \\
\midrule
Error Rate & 0.40\% ± 0.56\% & 6.67\% ± 0.46\% & 0.80\% ± 0.01\% & 0.00\% ± 0.00\% & 844 \\
\bottomrule
\end{tabular}}
\vspace{-4pt}
\caption{\footnotesize{Error types in semantic annotations across 100 sampled objects (844 outputs).}}
\label{tab:hallucination_appendix}
\vspace{-8pt}
\end{table}

As shown in Table~\ref{tab:hallucination_appendix}, GPT-4o maintains a low overall hallucination rate, with most errors falling into the category of semantically unreasonable but still structurally valid suggestions. Notably, there were no format violations, validating its strong instruction-following reliability for structured annotation generation.

\subsection{Expanded Human Verification of Annotations}

To further validate the annotation quality, we expanded our human verification from 50 to 200 objects and introduced more granular metrics. Each object was independently verified by three annotators.

\begin{table}[H]
\centering
\renewcommand\arraystretch{1.1}
\tabcolsep3.5pt
\resizebox{1.0\linewidth}{!}{
\begin{tabular}{c | c c c | c | c c | c}
\toprule
\textbf{Metric} & \multicolumn{4}{c|}{\textbf{Accuracy}} & \multicolumn{2}{c|}{\textbf{Completeness}} & \textbf{Diversity} \\
\cmidrule(r){2-8}
 & Keypoint & Direction & Consistency & Object-level* & Positional & Functional & Usage \\
\midrule
Success Rate & 98.70\% ± 0.45\% & 98.39\% ± 0.59\% & 92.69\% ± 1.15\% & 72.28\% ± 2.45\% & 92.52\% ± 3.43\% & 97.00\% ± 0.04\% & 97.00\% ± 0.04\% \\
\bottomrule
\end{tabular}}
\caption{\footnotesize{Expanded human verification results on 200 objects. Object-level accuracy considers a failure in any sub-metric as incorrect.}}
\label{tab:human_verification_appendix}
\vspace{-8pt}
\end{table}

These results confirm the annotation pipeline's effectiveness and robustness, supporting the integration of PASG-generated annotations in real manipulation settings.

\section{Author Contribution}
\noindent\textbf{Code Implementation} 
Zhihao Zhu implemented the foundational PASG framework and the automated pipeline. SiYu Pan built upon this with extensive optimizations, including prompt optimization and updates to the recognition backbone, helping to refine the automated pipeline. Yifan Zheng completed the benchmark design, model deployment, and testing code, as well as fine-tuning models. SiYu Pan and Zhihao Zhu implemented the adaptation of PASG in simulation scenarios and constructed the testing pipeline code. The code of Robotwin, SoM, Robocasa and LLaMA-Factory accelerated the implementation.

\noindent\textbf{Paper writing} 
Zhihao Zhu and Yao Mu finished introduction and methodology sections of the paper. Yifan Zheng and Zhihao Zhu wrote the experiments section. Zhihao Zhu provided all the visualizations shown in the paper. Yifan Zheng and Zhihao Zhu added results and analysis for their corresponding parts. Yao Mu carefully reviewed and revised the paper and gave feedback. Other authors help proofread and provide feedbacks.

\noindent\textbf{Experiments} 
Yifan Zheng led the design and construction of the benchmark, and completed experiments on model fine-tuning and performance evaluation. Zhihao Zhu and SiYu Pan co-led the performance evaluation experiments in simulation scenarios, and also conducted experiments on analysis and visualization.

\noindent Yao Mu is the main advisor of this project.

\section{Limitation}
While advantageous, PASG also has limitations. First, semantic annotations may inherit hallucinations from GPT-4o, despite our multi-round iterative refinement strategy. This process incurs significant computational costs, motivating future exploration of lightweight verification mechanisms (e.g., human-in-the-loop validation or physics-guided pruning) to enhance efficiency. Second, our framework mainly focuses on rigid objects, whereas deformable or articulated objects (e.g., hinged tools) also exhibit critical interaction primitives. Extending PASG requires redefining and adapting primitive categories to accommodate dynamic shape variations. This remains an open challenge but aligns with our ongoing efforts to broaden applicability.

\end{document}